\documentclass[lettersize,journal]{IEEEtran}

\usepackage{stfloats}

\usepackage{graphicx} 
\usepackage{xcolor}
\usepackage{amsmath}
\usepackage{amssymb}

\usepackage{multirow}
\usepackage{graphicx}
\usepackage{tcolorbox}

\usepackage{subfigure}
\usepackage{subcaption}
\usepackage{hyperref} 

\usepackage{algorithmicx}
\usepackage[ruled, vlined, linesnumbered, resetcount]{algorithm2e}
\SetKwInput{KwInput}{Input}                
\SetKwInput{KwOutput}{Output} 
\let\oldnl\nl
\newcommand{\nonl}{\renewcommand{\nl}{\let\nl\oldnl}}
\SetKwRepeat{Do}{do}{while}%
\def\HiLi{\leavevmode\rlap{\hbox to \hsize{\color{gray!35}\leaders\hrule height .8\baselineskip depth .5ex\hfill}}}
\def\HiLii{\leavevmode\rlap{\hbox to \hsize{\color{blue!10}\leaders\hrule height .8\baselineskip depth .5ex\hfill}}}

\begin{document}

\title{CAFD: Concept-Aware DNN Fault Detection using VLMs}

\author{Amin Abbasishahkoo,
        Mahboubeh Dadkhah,
        Lionel Briand,~\IEEEmembership{Fellow,~IEEE}
\thanks{Amin Abbasishahkoo is with the School of EECS, University of Ottawa,
Ottawa, ON K1N 6N5, Canada (e-mail: aabba038@uottawa.ca).}
\thanks{Mahboubeh Dadkhah is with the School of EECS, University of Ottawa,
Ottawa, ON K1N 6N5, Canada (e-mail: mdadkhah@uottawa.ca).}
\thanks{Lionel Briand is with the School of EECS, University of Ottawa, Ottawa, ON K1N 6N5, Canada, and also with the Research Ireland Lero centre for software, University of Limerick, Ireland (e-mail: lbriand@uottawa.ca).}

\thanks{Manuscript received March 1, 2025.}}

\maketitle
\begin{abstract}

Fault detection for Deep Neural Networks (DNNs) has received increasing attention in recent years. While more advanced hybrid approaches have been proposed to combine multiple sources of information and outperform earlier techniques, they often incur substantial computational overhead, limiting scalability and practicality in real-world settings.
In this paper, we introduce Concept-Aware Fault Detection (CAFD), a learning-based approach that achieves superior fault detection performance by effectively integrating multiple information sources while maintaining practical efficiency. Specifically, CAFD is trained using a carefully selected set of informative features, including model-based signals derived from the DNN’s outputs (e.g., output probabilities, logit layer activations, and uncertainty metrics), distance-based features, and a novel concept-based feature.
We introduce the Concept Failure Ratio (CFR) as a novel concept-based feature. CFR captures the semantic characteristics of inputs by leveraging Vision-Language Models (VLMs) to extract textual concepts from images and quantifying the likelihood that the presence of each concept is associated with DNN failures. By incorporating CFR alongside model-based and distance-based features, CAFD benefits from complementary semantic information, enabling more effective fault detection.
Our results demonstrate that CFR provides complementary semantic insight and serves as an effective indicator for DNN fault detection. 
We conduct an extensive empirical evaluation of CAFD, comparing it against five state-of-the-art baselines across three subject DNN models and datasets, including ImageNet. Across a wide range of constrained selection budgets, CAFD consistently outperforms all baselines in Fault Detection Rate (FDR), particularly under highly constrained budgets. Specifically, CAFD achieves average FDR improvements of 18.3\% across all investigated subjects and budget sizes.

\end{abstract}

\IEEEpeerreviewmaketitle

%


\section{Introduction}
\label{sec:Introduction}

Deep Neural Networks (DNNs) have been widely adopted across various application domains, including computer vision, healthcare, and software engineering ~\cite{chai2021deep, yang2022survey, shahkoo2023autonomous}. As with traditional software systems, DNNs must be thoroughly tested to ensure reliability, robustness, and correctness, particularly in safety-critical domains. However, effective testing of DNNs requires a diverse set of labeled inputs to identify a wide range of distinct faults. In practice, manually labeling such inputs is often impractical due to the significant cost and human effort involved, particularly in domains that require specialized expertise. This challenge is further exacerbated when dealing with large datasets, where they often contain many inputs that reveal redundant faults, resulting in inefficient use of limited labeling resources. To address this limitation, we propose Concept-Aware Fault Detection (CAFD), a learning-based approach that achieves superior performance in detecting diverse faults while remaining practically efficient, even on large datasets like ImageNet. CAFD is a fault detection model trained with an effective set of input features, including representative features from the DNN's output, uncertainty-based and distance-based metrics, and a novel concept-based metric derived from Vision-Language Models (VLMs).

In the context of DNN testing, several fault detection approaches have been proposed in recent years~\cite{hu2024test}. 
Among the most widely used methods are uncertainty-based approaches, which estimate the confidence of the model under test (MUT) on unlabeled inputs. These approaches can be categorized into probability-based uncertainty metrics, which solely rely on the MUT’s output probability distribution~\cite{feng2020deepgini, hu2024test, Weiss2022SimpleTechniques}, and neighbor-aware uncertainty metrics, which additionally incorporate information from the input’s nearest neighbors~\cite{li2024distance, bao2023defense}. While neighbor-aware uncertainty metrics have shown improved Fault Detection Rate (FDR) in some cases~\cite{li2024distance, abbasishahkoo2025metasel}, they are less efficient than probability-based metrics, particularly on larger datasets. In addition, diversity is often considered a key metric in fault detection, as it promotes selecting inputs with diverse characteristics, enabling broader coverage of the input space and increasing the likelihood of detecting a wide range of distinct faults~\cite{gao2022adaptive}.

Hybrid and learning-based approaches aim to leverage the complementary strengths of multiple information sources to enhance the effectiveness of DNN fault detection~\cite{aghababaeyan2024deepgd, sun2023robust, li2021testrank, demir2024test}. Several of these approaches integrate both uncertainty and diversity~\cite{aghababaeyan2024deepgd, sun2023robust}, aiming to select inputs that cover different regions of the input space to reduce redundancy while maximizing the likelihood of detecting faults. However, leveraging multiple metrics typically incurs additional computational overhead. In some cases, this overhead becomes substantial enough to limit the feasibility and practical applicability of such approaches~\cite{aghababaeyan2024deepgd, abbasishahkoo2025metasel}. Consequently, carefully selecting and combining the most informative yet efficient input features is crucial for training a fault detection model that achieves an effective trade-off between performance and efficiency. To achieve this goal, 
we conducted an extensive exploration, investigating various features to identify the most effective yet practically efficient combination.

Furthermore, we introduce the Concept Failure Ratio (CFR), a novel metric derived from VLMs. Specifically, we employ CLIP (Contrastive Language–Image Pretraining)~\cite{radford2021learning}, a widely used VLM consisting of an image encoder and a text encoder that map inputs into a shared embedding space. This shared representation enables us to measure the similarity between each image input and a set of candidate textual concepts and extract highly similar concepts that represent that input. CFR is a metric that quantifies, for each concept, the likelihood that its presence in a given input leads to MUT failing to predict that input. To construct a set of candidate textual concepts for each DNN and to calculate the CFR score for each concept, we use the MUT's training set. 
Subsequently, for each unlabeled input, we identify its top-$m$ most similar concepts and use their corresponding CFR scores to estimate the likelihood that the input can detect faults in the DNN.
Our results demonstrate that, when combined with information from the DNN's output, uncertainty metrics, and distance-based features, CFR provides complementary semantic information and serves as an effective indicator for DNN fault detection. To construct CAFD, we investigated several learning models, including classical machine learning models and Neural Networks (NN), and ultimately selected Logistic Regression (LR) because it delivered strong performance while maintaining high computational efficiency.




We conduct a comprehensive empirical evaluation across multiple datasets, including large-scale datasets such as ImageNet~\cite{ILSVRC15}, and compare CAFD with five state-of-the-art (SOTA) baselines. The results demonstrate that CAFD consistently outperforms existing methods in detecting faults across all subjects and selection budgets. Particularly under highly constrained budget sizes (1\%, 3\%, and 5\%), CAFD achieves average FDR improvements of 10\%-19\% over the second-best-performing approach across the evaluated datasets. Furthermore, the second-best approach varies across datasets and budget sizes, confirming CAFD as the only reliable approach. In addition to its superior performance, our evaluation results demonstrate that CAFD remains practically efficient even for larger datasets, requiring around 840 seconds to rank the entire ImageNet test set.

The key contributions of this paper are as follows:
\begin{itemize}
    \item We introduce CAFD, a novel learning-based approach that achieves superior fault-detection performance while remaining practically efficient. As a learning-based approach, CAFD is trained using an effective set of input features, including the DNN model's output, uncertainty-based and distance-based features, and a novel concept-based metric extracted from image inputs using VLMs. 
    

    \item We introduce CFR, a novel metric that estimates, for each textual concept extracted from an input image, the likelihood that its presence is associated with the MUT's failure. To extract these textual concepts, we leverage VLMs. The resulting CFR scores are leveraged as one of the features for training the CAFD model. 
    
    \item We conduct a comprehensive empirical evaluation on three subject DNN models and datasets, including larger datasets such as ImageNet, demonstrating that CAFD consistently outperforms SOTA baselines across different selection budgets in terms of FDR. Moreover, our results demonstrate that the second-best baseline greatly varies across datasets and budget sizes, highlighting CAFD as the only approach that consistently delivers superior performance.
    
    \item We empirically evaluate CAFD's computational efficiency and demonstrate that although it is not the most efficient among the investigated approaches, it remains practically efficient even on larger datasets such as ImageNet. These findings confirm that CAFD uniquely combines consistent, superior fault detection performance with computational practicality.
    
\end{itemize}



The remainder of this paper is organized as follows: Section II provides background, including a brief overview of the fault detection baselines used in our experiments, a summary of concept extraction from images using VLMs, and fault estimation in DNN models. Section III introduces the CAFD model and its training feature set, including the CFR score. Section IV describes the experimental design used in our evaluations. Section V reports and discusses the results for each research question. Section VI reviews related work, and Section VII concludes the paper.

\section{Background}
\label{sec:Background}

This section provides an overview of SOTA fault-detection approaches for DNNs, which serve as baselines in our study.
Additionally, by introducing a concept-aware learning-based method for detecting fault-revealing inputs using VLMs, we explain how they can be used to extract concepts from images. We also discuss the method we used to estimate faults in DNNs, which we used to evaluate our approach.

\subsection{Fault Detection in DNNs}
In recent years, numerous approaches have been proposed for DNN fault detection. However, to ensure our experiments remain computationally feasible, we intentionally exclude certain existing approaches based on their limited effectiveness or high computational cost, as reported by prior works. In particular, we exclude approaches that have been 
outperformed by some of the baselines included in our experiments. 
Approaches such as Surprise Adequacy (SA) metrics, proposed by Kim \textit{et al.}~\cite{kim2019guiding}, Adaptive Test Selection (ATS)~\cite{gao2022adaptive}, and Neuron Coverage (NC) metrics ~\cite{pei2017deepxplore, Ma2018DeepGaugeMT}, have been excluded due to their limited effectiveness~\cite{gao2022adaptive, hu2024test, aghababaeyan2024deepgd, sun2023robust}. SA metrics measure how surprising a given input is for the MUT relative to its training set, while NC evaluates neuron-activation coverage.
Similarly, ATS, introduced by Gao \textit{et al.}~\cite{gao2022adaptive}, has consistently been outperformed by more recent baselines~\cite{aghababaeyan2024deepgd, hu2024test}. Moreover, ATS iteratively selects inputs based on the dissimilarity between each remaining input and the currently selected subset, making it one of the most computationally intensive approaches~\cite{hu2024test}.

Mutation-based approaches are also excluded from this study due to their high computational cost and even impracticality in many cases~\cite{abbasishahkoo2024teasma}. 
These approaches are highly time-consuming because they require generating multiple mutants and executing each unlabeled test input against all mutants~\cite{abbasishahkoo2024teasma, wang2021prioritizing}.
Furthermore, we excluded 
DeepGD proposed by Aghababaeyan \textit{et al.}~\cite{aghababaeyan2024deepgd}, which is a hybrid search-based approach, leveraging the Non-dominated Sorting Genetic Algorithm (NSGA-II)~\cite{deb2002fast}, for fault detection that combines uncertainty with geometric diversity.
The scalability challenges of applying DeepGD have been empirically highlighted in a recent study~\cite{abbasishahkoo2026highly}, 
which reports a substantial decline in efficiency when these techniques were applied to large-scale datasets such as ImageNet. 


Among fault-detection approaches for DNNs, uncertainty-based methods constitute an important baseline category. The primary objective of these approaches is to identify inputs for which the MUT is less confident in its predictions~\cite{feng2020deepgini, hu2024test, Weiss2022SimpleTechniques}. As a result, these methods tend to select inputs that lie close to the MUT's decision boundary. 
Uncertainty-based approaches are not only highly efficient, but several of them have also proven strongly effective in detecting faults in DNN models~\cite{hu2024test, Weiss2022SimpleTechniques, feng2020deepgini, li2024distance, Weiss2022SimpleTechniques}. 
Uncertainty metrics measure the DNN model's confidence in predicting a given input, 
by analyzing the model's output probability distribution~\cite{hu2024test} for that input or by combining it with information from the input's nearest neighbors within the dataset. DeepGini~\cite{feng2020deepgini}, Margin~\cite{hu2024test}, and Vanilla Softmax~\cite{Weiss2022SimpleTechniques} are well-established uncertainty metrics that rely solely on the model's predicted probability for a given input. 
We include all of these three metrics in our experiments. 

   
         

DeepGini~\cite{feng2020deepgini} is a widely recognized metric that 
has been extensively used as a baseline in DNN testing research~\cite{aghababaeyan2024deepgd, gao2022adaptive, dang2023graphprior, hu2022empirical, wang2021prioritizing}.
DeepGini selects inputs with a larger spread of softmax values across classes since they reflect a higher level of classification uncertainty, and is defined as $1 - \sum_{i=1}^{C} P_i(x)^2$, where $P_i(x)$ is the predicted probability of class $C_i$.
The Vanilla Softmax metric~\cite{hu2023evaluating} is computed as one minus the maximum activation value in the output softmax layer ($1 -  \max_{i=1}^{C} {P_i(x)}$), providing a straightforward measure of uncertainty. 
With minimal computational and theoretical complexity, both serve as simple yet effective baselines for DNN fault detection. 
The Margin score measures the difference between the two highest predicted probabilities in the MUT's output probability vector and is defined as $p_m(x) - p_n(x)$. 
Each of these uncertainty metrics has been shown to be effective in prior studies~\cite{aghababaeyan2024deepgd, li2024distance, Weiss2022SimpleTechniques}, though their performance varies across different settings and assumptions.
Given these mixed findings and the complementary insights from each metric, we include all three uncertainty metrics---DeepGini, Vanilla Softmax, and Margin score---as baselines in our experiments to ensure a comprehensive evaluation.

Recent studies have explored leveraging the available information from a given input’s neighbors to estimate the model’s uncertainty in its prediction for that input~\cite{li2024distance, bao2023defense}. One of the most effective methods in this category is DATIS (Distance-Aware Test Input Selection), proposed by Li \textit{et al.}~\cite{li2024distance}, which has been shown to outperform prior input selection techniques in DNN fault detection~\cite{abbasishahkoo2025metasel, li2024distance}. Unlike uncertainty metrics that rely solely on predicted probabilities, DATIS estimates uncertainty based on the ground-truth labels of the input’s nearest neighbors within the MUT's labeled training set. First, it calculates the support of the training inputs for the DNN's prediction for a given input $x$ by calculating the NED~\cite{karpusha2020calibrated} score for each output class $c$ (i.e., $p^*_c (x)$). Then, the DATIS uncertainty score for input $x$ is computed as $\text{DATIS}(x) = p^*_n / p^*_m$, where $m$ is the class predicted by the DNN model for input $x$, $p^*_m$ denotes the most supported prediction distinct from the DNN predicted class $m$. Therefore, a higher value of $\text{DATIS}(x)$ indicates that the training inputs weakly support the MUT’s prediction for $x$.


Hybrid input selection approaches aim to leverage the complementary strengths of multiple metrics to enhance the effectiveness of DNN fault detection. 
Robust Test Selection (RTS) proposed by Sun \textit{et al.}~\cite{sun2023robust} is a hybrid test input selection method that integrates both uncertainty and diversity for DNN fault detection. RTS partitions unlabeled candidate inputs into three categories---noise, suspicious, and successful (correctly classified)---based on the MUT’s output probability distributions. It then selects inputs sequentially according to a fixed priority: suspicious inputs, then successful inputs, and finally noise inputs. Within each category, RTS prioritizes inputs using a novel probability-tier-matrix-based metric. As a result, RTS not only focuses on the diversity of the selected inputs but also avoids prioritizing inputs unlikely to reveal faults in the MUT.
Their comprehensive evaluations demonstrate RTS's superior effectiveness in fault detection compared to both NC~\cite{pei2017deepxplore, Ma2018DeepGaugeMT} and SA~\cite{kim2023evaluating} metrics selection approaches~\cite{sun2023robust}.



\subsection{Concept Extraction using VLMs}
\label{sec:concept_extraction}
Recent studies have demonstrated the promising advantages of leveraging semantic concepts extracted from image inputs~\cite{abbasishahkoo2026highly, hu2025debugging, mangal2024concept, moayeri2023text}. This progress has been largely driven by recent advances in VLMs, a new category of models that learn joint representations from both visual and textual inputs. Unlike traditional vision models that rely on manually annotated datasets with fixed labels, VLMs are trained on massive collections of image-text pairs. This large-scale, multimodal training enables them to generalize across diverse tasks and domains, more importantly, to represent and describe visual concepts in natural language.

One of the most well-known VLMs is CLIP~\cite{radford2021learning}, proposed by OpenAI. CLIP consists of two jointly trained encoders: an image encoder, such as ResNet or Vision Transformer (ViT), and a Transformer-based text encoder. Each encoder projects its input into a shared multimodal embedding space. 
Recent studies have increasingly relied on CLIP for extracting semantic concepts from image inputs. In particular, Abbasishahkoo \textit{et al.}~\cite{abbasishahkoo2026highly} leveraged CLIP to extract highly relevant concepts for image inputs. In their approach, images are embedded using CLIP’s image encoder into CLIP space, 
and candidate concepts are represented as natural-language prompts embedded by the text encoder. Concept candidates are constructed from a large-scale knowledge base such as Visual Genome~\cite{krishna2017visual}, which contains a diverse set of related concepts in the underlying vision-based tasks. A zero-shot classification procedure is then performed by comparing each image embedding with a set of concept embeddings derived from the knowledge base, and the top-$m$ concepts with the highest similarity scores are selected to represent each image ($m=10$ in their original work), and the union of these concepts forms the Representative Concept Set (RCS). This RCS is subsequently used to extract semantic concepts for unseen test inputs.

To extract related concepts from an input image using CLIP, each image must be processed by CLIP’s image encoder to obtain its corresponding embedding in the shared embedding space. Consequently, in test input selection scenarios involving a large set of unlabeled inputs, CLIP must be executed once per input, which introduces significant computational overhead when scaling to large datasets. To efficiently extract concepts, Abbasishahkoo \textit{et al.}.~\cite{abbasishahkoo2026highly} adopt a linear alignment strategy originally proposed by Moayeri \textit{et al.}~\cite{moayeri2023text}, which maps DNN latent representations directly into the CLIP embedding space. Consequently, it eliminates the need to repeatedly run CLIP during inference and significantly reduces computational overhead.
Following their pipeline, we extract semantic concepts from training inputs using the CLIP model, and use the Visual Genome knowledge base as the source of concepts, since our experiments are conducted on general image datasets such as ImageNet~\cite{abbasishahkoo2026highly}.
Then, we construct the RCS for each dataset and train a linear aligner for each DNN model to efficiently extract relevant concepts for unlabeled test inputs. We use the trained aligner to map the DNN representations of image inputs into the CLIP space and compute their similarity with the precomputed RCS embeddings to retrieve the top-$m$ ($m=10$) most relevant concepts.

\subsection{Fault Estimation}
The number of failure-causing test inputs (i.e., mispredictions) is a commonly used metric for evaluating DNN testing approaches~\cite{pei2017deepxplore, kim2019guiding, feng2020deepgini}. However, the number of failures does not correctly reflect the actual number of underlying faults. Indeed, a single fault within a DNN model can cause multiple test inputs to fail. Subsequently, a test set that identifies a large number of failures may actually detect only a few underlying faults, whereas another test set with the same number of failures could detect more distinct faults. Therefore, using failure rates to compare test input selection techniques in terms of their ability to detect faults is highly misleading, as many test inputs may fail for the same reasons~\cite{aghababaeyan2023black, fahmy2021supporting}. 

Consequently, we rely on a recent approach proposed by Aghababaeyan \textit{et al.}~\cite{aghababaeyan2023black} to estimate faults in a DNN model, where faults are defined as distinct root causes of failures~\cite{fahmy2021supporting}. Aghababaeyan \textit{et al.}~\cite{aghababaeyan2023black} cluster failure-causing inputs with similar characteristics through three main steps: feature extraction, dimensionality reduction, and density-based clustering. First, VGG-16~\cite{simonyan2014very} is used to extract a feature matrix from the failure inputs of the training and test sets, which serves as the basis for clustering. Then, dimensionality reduction is applied to improve clustering performance in the high-dimensional feature space. Finally, the HDBSCAN algorithm~\cite{campello2013density} is employed to cluster the failure inputs based on the extracted features.

Aghababaeyan \textit{et al.}~\cite{aghababaeyan2023black} relied on both manual and metric-based evaluation to identify the best hyperparameter configuration from multiple clustering settings. For the metric-based evaluation, they used two established metrics:  Silhouette score~\cite{rousseeuw1987silhouettes} and Density-Based Clustering Validation (DBCV)~\cite{moulavi2014density}. 
To complement these quantitative assessments, they conducted a manual analysis of the quality of the final selected clusters. Specifically, they generated heatmaps for each cluster to visualize the distribution of features across inputs within the cluster. Their empirical observations indicated that inputs within the same cluster lead to DNN failures stemming from the same underlying root cause, whereas inputs across different clusters cause failures stemming from distinct root causes. Therefore, if a test set includes at least one input from a cluster, it can detect the underlying fault associated with that cluster.

Consequently, in this study, we rely on the clustering approach introduced by Aghababaeyan \textit{et al.}~\cite{aghababaeyan2023black} and adopt the same manual and metric-based approach to assess the identified fault clusters.  In our experiments, we used the original implementation and parameter settings provided by Aghababaeyan \textit{et al.}. Table~\ref{tab:Subjects} reports the total number of faults identified for each subject.

\begin{figure*}
    \includegraphics[width = 0.9\textwidth]{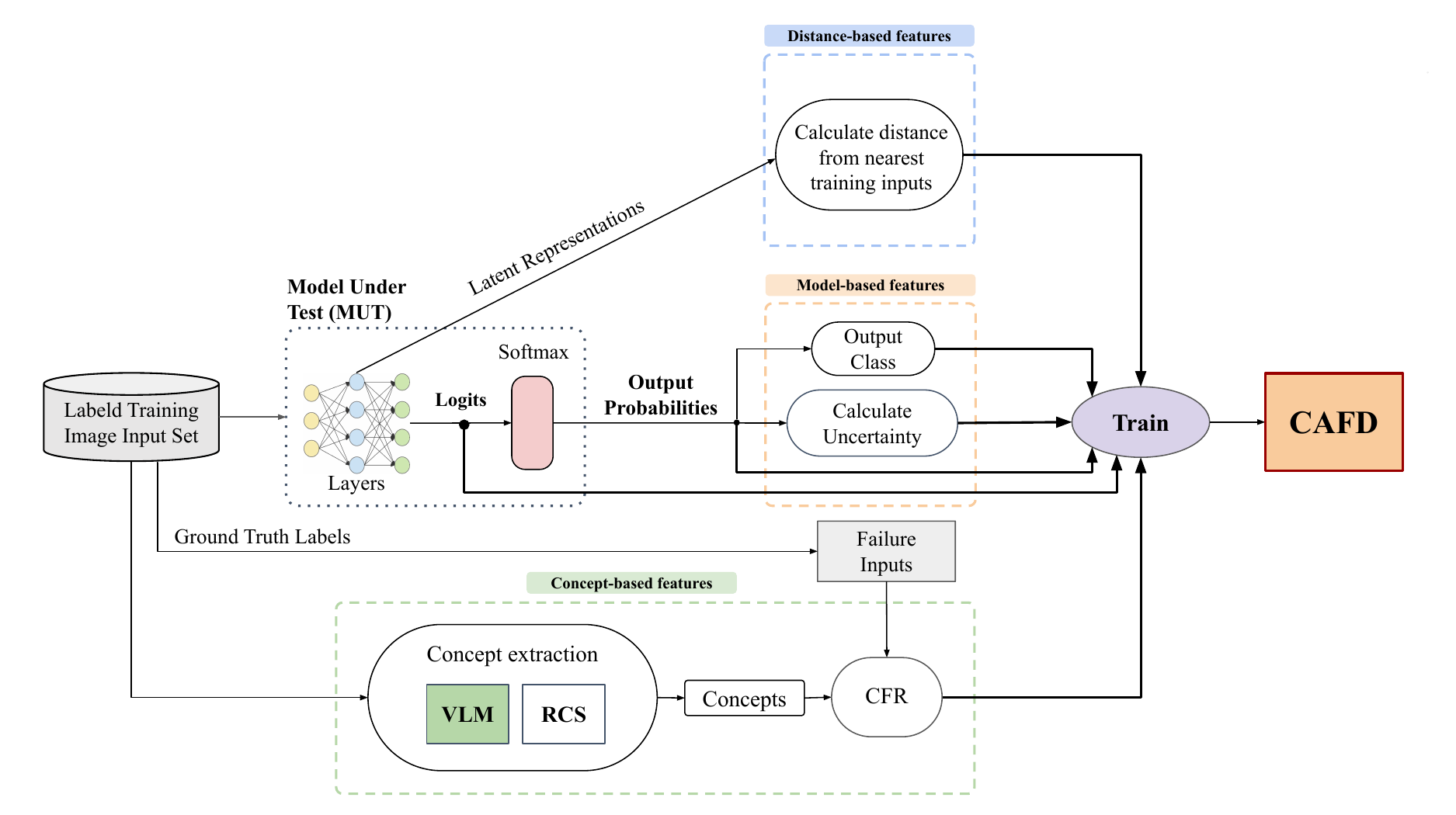}
    \centering
    \caption{The process of training CAFD}
    \label{fig:CAFD}
\end{figure*}

\section{Approach}
\label{sec:Approach}

In this section, we introduce Concept-Aware Fault Detection (CAFD), a novel learning-based fault detection approach for DNN models that leverages VLMs. 
CAFD aims to identify unlabeled inputs that are more likely to reveal faults in the MUT.
As a learning-based approach, designing CAFD involves selecting a representative and informative set of input features for training, constructing an effective training set, and selecting an appropriate model architecture. In the following, we elaborate on each of these components in detail.

\subsection{Feature Selection}
One of the main challenges in designing learning-based approaches is selecting an effective set of features that yields superior model performance. This task is inherently complex, as different combinations of features can significantly influence the overall model behavior. Therefore, selecting a compact yet informative feature set is essential to achieving a superior fault-detection rate while maintaining efficiency. 


To address this challenge, inspired by the demonstrated effectiveness of prominent DNN fault detection methods~\cite{abbasishahkoo2025metasel, abbasishahkoo2026highly, li2024distance}, we selected representative inputs to construct an informative feature set. 
Specifically,  as shown in Figure~\ref{fig:CAFD}, we included features from three complementary categories: 
(1) model-based features, derived from the output of the MUT, including the MUT's output probabilities, logit layer outputs, and uncertainty scores; 
(2) distance-based features, obtained from distance relationships between inputs; and (3) concept-based features, extracted from image inputs using VLMs, which capture semantic characteristics of inputs through their concepts.

While model-based and distance-based features are well-established and have been leveraged in prior learning-based approaches~\cite{abbasishahkoo2025metasel, li2021testrank, demir2024test}, the extraction of underlying semantic concepts from inputs has only recently become feasible due to advances in VLMs. Recent studies have demonstrated the effectiveness of such concepts in capturing diversity in image inputs~\cite{abbasishahkoo2026highly, hu2025debugging}. Motivated by this, we investigate the use of concept-level information to guide fault detection. We hypothesize that certain semantic concepts are more strongly associated with failures of the MUT. Our preliminary study provides empirical support for this hypothesis, indicating that inputs with specific concepts are more likely to trigger failures in the MUT. 
Based on this observation, we introduce a concept-based feature for training our fault detection model.
We demonstrate that incorporating this feature alongside model-based and distance-based features provides complementary semantic insights and enables the resulting model to achieve a superior FDR.
To identify an effective feature set for training a model with a high fault-detection rate, we conducted a preliminary study using a limited set of input datasets and models to save time and resources. We explored various combinations of inputs across the three categories and trained a diverse set of candidate models, including simpler DNNs and classical machine learning models such as logistic regression and decision trees. We then evaluated the resulting fault detection models in terms of FDR on the selected datasets. Based on this analysis, we selected the feature combination that consistently achieved the highest FDR across all considered settings.
In the following, we discuss the final set of selected features in detail.

\subsubsection{Model-based features}
This category encompasses features derived from the MUT's prediction for a given input. In addition to the MUT's predicted output class, this category includes output probabilities, the logit layer's output, and uncertainty scores. 
The output probabilities are derived from the softmax layer's output and capture the MUT’s prediction distribution across all output classes. Inputs with similar probabilities across all classes are shown to have a higher likelihood of failure~\cite{feng2020deepgini}. 
In contrast, the logit layer outputs represent the raw, unnormalized outputs of the MUT before the softmax layer. Unlike output probabilities, logits preserve the relative differences between the MUT's predictions for different classes, providing a more informative and fine-grained view of the model’s decision-making. As shown in ~\cite{abbasishahkoo2025metasel}, this feature enables the model to capture subtle distinctions in the logit-layer output distributions that are not preserved in probability outputs due to softmax normalization, thereby helping identify distinct failure-prone inputs.



Uncertainty scores derived from the MUT's output probabilities have also been widely used in DNN failure detection studies ~\cite{feng2020deepgini, hu2024test} and have been shown to be effective for failure detection in DNNs. Although these uncertainty scores are derived from output probabilities and combining these inputs seems redundant, our results show that explicitly incorporating them reduces the need for the prediction model to infer complex patterns from raw information and improves failure detection. 
Several metrics have been proposed to quantify uncertainty based on the MUT's output~\cite{feng2020deepgini, hu2024test, Weiss2022SimpleTechniques}. We investigated three simple metrics, including DeepGini, Margin, and Vanilla, due to their demonstrated efficiency and effectiveness~\cite{Weiss2022SimpleTechniques}.

In our preliminary study, we investigated the effectiveness of different combinations of model-based features, including output probabilities, the logit layer output, uncertainty scores, and the MUT's predicted output class. Specifically, we evaluated feature sets containing different subsets of these model-based features along with features from the other two categories, ranging from pairwise combinations to the inclusion of all features. Our results on a limited set of datasets showed that the most effective feature set, which consistently leads the model to achieve superior FDR performance, includes all four features. We observed that excluding any of these features results in noticeable performance degradation. Consequently, we incorporate all four model-based features into the final feature set used for constructing CAFD. We also observed that among the three investigated output probability-based uncertainty metrics, DeepGini consistently achieves the best performance, while Margin and Vanilla exhibit comparable but lower effectiveness.



\subsubsection{Distance-based features}



The distances between representations of inputs in a latent space have been extensively studied in prior work for a variety of purposes, including calculating surprise adequacy scores~\cite{kim2023evaluating}, measuring diversity and guiding adaptive selection strategies~\cite{gao2022adaptive}, and estimating the MUT's uncertainty using neighbor-aware uncertainty metrics. 
Specifically, we leveraged the NED score~\cite{karpusha2020calibrated}, a weighted score based on the distances of the given input from its nearest training neighbors. This score estimates the support of neighbors for the MUT's prediction on that input, reflecting the level of agreement between the model’s output and the ground-truth labels of the nearest training samples. The NED score has also been adopted by DATIS~\cite{li2024distance} to estimate the MUT's uncertainty and guide DNN fault detection, achieving strong fault-detection performance. 
In particular, for a given input $t$, the NED score is defined as:

\begin{equation}
P(t) = \frac{\sum_{i=1}^{K} \exp\left(-\|Z_t - Z_i\|_2^2 / \tau\right) \mathbb{I}(\{Y_i = \hat{Y}_t)}{\sum_{i=1}^{K} \exp\left(-\|Z_t - Z_i\|_2^2 / \tau\right)}
\end{equation}

\noindent where $Y_i$ denotes the ground-truth label of the $i$-th neighbor and $\hat{Y}_t$ is the label predicted by the DNN for input $t$. We extract latent representations from the penultimate layer of the MUT, denoted by $Z_t$ for the input and $Z_i$ for its $K$ nearest training samples. The indicator function $\mathbb{I}(\{Y_i = \hat{Y}_t)$ equals 1 when the neighbor’s actual label matches the predicted label of input $t$, and 0 otherwise. The parameter $\tau > 0$ controls the influence of distance in the exponential weighting. In our experiments, we use the default value of $\tau$ suggested in the original NED paper~\cite{karpusha2020calibrated}.

Including this score in CAFD's feature set enables us to identify inputs with low support from their training neighbors, indicating limited local agreement with MUT’s prediction. Such inputs often correspond to regions of lower confidence in the MUT's input space, either because neighboring inputs exhibit conflicting labels or because the input is relatively isolated. As a result, these inputs tend to be more diverse with respect to the training inputs and are more likely to detect distinct faults.

\subsubsection{Concept-based features}
\label{sec:cfr}




Unlike model-based and distance-based features, which primarily capture MUT's predictions and the spatial characteristics of inputs, we introduce a concept-based feature that incorporates the semantic characteristics of inputs by representing them with concepts extracted using a VLM. 
In particular, we propose \textit{Concept Failure Ratio} (CFR),  a 
score that, for a given concept, measures the likelihood of its presence in an input leading to the MUT's failure, thus serving as an informative indicator for identifying inputs that are likely to detect MUT's failures. 





To calculate this metric, we first construct a set of Representative Concept Set (RCS) for the MUT based on its training set using a VLM, following the procedure previously described in Section~\ref{sec:concept_extraction}. RCS includes all unique concepts extracted from images in MUT's training set. 
We then compute the CFR for each concept in the RCS by measuring how frequently it appears in failure-causing inputs within the MUT's training set.
Let $Con_i$ denote a concept from the RCS. The CFR for concept $Con_i$ is defined as:

\begin{equation}
CFR(Con_i) = \frac{\#FaultyInputs(Con_i)}{\#TotalInputs(Con_i)}
\end{equation}

\noindent where $\#FaultyInputs(Con_i)$ denotes the number of MUT's training inputs containing concept $Con_i$ that lead to MUT's failure, and $\#TotalInputs(Con_i)$ denotes the total number of training inputs containing concept $Con_i$. A higher $CFR(Con_i)$ value indicates that 
inputs containing concept $Con_i$ are more likely to trigger failures in the MUT. Conversely, a lower value indicates that the concept is less associated with the failure in the MUT, and inputs containing it are more likely to be correctly predicted by the MUT. 
Next, for a given input $t$, we construct a concept-level vector by aggregating the \textit{CFR} values of its top-$m$ most similar concepts. Each CFR value reflects the likelihood that the input can detect a fault in the MUT with respect to a specific extracted concept. By integrating the CFR values of highly related concepts, CAFD captures complementary semantic information, enabling more accurate estimation of the input’s fault-detection potential. Specifically, for an input $t$ we extract $m$ concepts with highest similarity $\{Con_1, Con_2, \dots, Con_m\}$, as described in Section~\ref{sec:concept_extraction}. We then form a vector comprising the corresponding CFR values, $\{\text{CFR}(Con_1), \text{CFR}(Con_2), \dots, \text{CFR}(Con_m)\}$, which is the final input feature selected to construct CAFD. Our analysis on the contribution of selected features in constructing a superior fault detection model demonstrates that this concept-based feature provides complementary information to both model-based and distance-based features, and plays a critical role in enabling CAFD to achieve superior fault detection performance.


\subsubsection{Feature Importance Evaluation}


To further investigate the contribution of different features and exclude features that do not significantly contribute to the CAFD, we conducted a feature importance analysis using coefficient magnitude analysis~\cite{saarela2021comparison} in a logistic regression (LR) model. We used LR because, as we will describe in Section~\ref{sec:model}, CAFD trained with LR achieves superior fault-detection performance compared to other models, such as Random Forest (RF) and Support Vector Machines (SVM). 
Features with higher absolute estimated coefficients are considered more influential.
The results indicate that the selected concept-based CFR and distance-based features are the most influential on fault detection, followed by model-based features. 

Notably, we observed that incorporating similarity scores between inputs and their extracted concepts does not improve performance. This is likely because inputs that are distant in the input space may still exhibit similar similarity scores to their associated concepts, thereby reducing their discriminative power for the prediction model. Similarly, incorporating training support for all output classes 
does not lead to performance improvements. This is likely because these class-wise support values are largely shared across inputs and fail to capture input-specific characteristics, thereby degrading the model’s discriminative ability.
We also employed additional feature importance techniques, including Recursive Feature Elimination (RFE)~\cite{guyon2002gene} and Odds Ratios~\cite{szumilas2010explaining}, and we observed consistent results. RFE iteratively removes the least important features based on model-derived importance scores to identify the most relevant subset. Odds Ratios, derived from the LR model, quantify the strength of association between each feature and the target outcome, indicating how changes in feature values affect prediction likelihood.

\subsection{Model selection and training set}
\label{sec:model}

The core of CAFD is a learning model that estimates the likelihood that an unlabeled test input can detect a fault. In our preliminary study, we explored several learning models, ranging from simple to more complex, including LR, RF, SVM, and NN. Among these, LR consistently achieved strong performance while maintaining high computational efficiency. 
To train the CAFD model, we need a labeled training set in which inputs are labeled according to their fault detection ability with respect to the MUT. 
As a result, we use MUT's training inputs as ground truth labels, allowing us to determine whether each input leads to a correct prediction or indicates a failure in the MUT. In addition, for each input, we extract the corresponding feature values required for training CAFD, including selected model-based, distance-based, and concept-based features, as illustrated in Figure~\ref{fig:CAFD}. 





\section{Experiment Design}
\label{sec:ExperimentDesign}

In this section, we describe the experiments designed for an empirical evaluation of CAFD, including the research questions we address, the subjects on which we conduct the evaluation, the configuration for extracting concepts from image inputs using a VLM, and the evaluation metrics.

\subsection{Research Questions}
\label{sec:ResearchQuestions}
We conducted an extensive evaluation to answer two key questions regarding both the effectiveness and efficiency of CAFD, as follows:

\textbf{\textit{RQ1: How effective is CAFD in detecting faults?
}}
To answer this question, we investigate the effectiveness of our approach and compare it with the baseline approaches described in Section~\ref{sec:Background} in terms of their ability to reveal distinct faults in DNNs.
For this purpose, we compare the FDR of test subsets selected by CAFD and each baseline. We perform our evaluation across selection budgets of 1\%, 3\%, 5\%, 7\%, 10\%, and 12\% of the test set for each subject dataset and DNN model. We specifically focus on smaller selection budgets in our experiments to reflect real-world scenarios where labeling budgets are highly constrained.

\textbf{\textit{RQ2: How computationally efficient is CAFD?
}}

In this research question, we evaluate the efficiency of CAFD by comparing its total execution time with that of baseline approaches. This evaluation is especially important for assessing the scalability of DNN-based fault-detection approaches. In particular, for larger datasets, the excessively long execution times could potentially limit the practicality and real-world applicability of these approaches. For this purpose, we compute the execution time of our approach and the baselines on the entire test set and compare them. This is because each method requires a single execution on the entire test set to rank all inputs. Then the final test subset can be selected from the top of the ranked list based on the available budget size.



\begin{table}[b]
    \centering   
    \footnotesize 
    \caption{Information about the datasets and DNN models}
    \resizebox{\columnwidth}{!}{
    \begin{tabular}{|c|      c|  c|  c| c |  }
    \hline 
    \multirow{2}{*}{Dataset}  &\multirow{2}{*}{Model}   &Training set    &Test set &Total \\ 
       &                        &size    &size &faults   \\ \hline 
          Cifar-10  &ResNet18  &50,000   & 10,000  &319  \\  \hline
          Cifar-100  &ResNet152  &50,000   & 10,000 &690 \\  \hline
          ImageNet       &ResNet101  &1,281,167    &50,000 &5801             \\  \hline
    \end{tabular}
    }
    \label{tab:Subjects}
\end{table}

\subsection{Subjects}
\label{sec:Subjects}
We conducted our experiments using a collection of well-known image datasets, including Cifar-10~\cite{krizhevsky2009learning}, Cifar-100~\cite{krizhevsky2009learning}, and ImageNet~\cite{ILSVRC15} that have been utilized in numerous empirical studies on DNN testing~\cite{shen2020multiple, aghababaeyan2024deepgd, abbasishahkoo2024teasma, hu2022empirical, berend2020cats}. 

Cifar-10 is a widely used dataset of images from 10 categories, such as cats, dogs, airplanes, and cars. Cifar-100 is another widely used dataset that contains 100 classes and a total of 60,000 images. Both Cifar-10 and Cifar-100 are standard benchmarks in computer vision and consist of 32×32 color images. Each dataset is split into 50,000 training images and 10,000 test images.
Moreover, to evaluate the scalability of CAFD on larger datasets, we include ImageNet, a widely recognized large-scale dataset for visual object recognition research in computer vision. It contains over 14 million realistic, feature-rich, labeled images across 1,000 object categories~\cite{ILSVRC15}. We leverage its most commonly adopted subset, ImageNet-1k, from the ILSVRC2012 competition, which includes 1,281,167 training images and 50,000 test images for evaluation. The images in ImageNet are 224×224 pixels and have a higher resolution than the Cifar datasets.

Using these datasets, we trained three DNN models using ResNet18~\cite{he2016deep}, ResNet152~\cite{he2016deep}, and ResNet101~\cite{he2016deep}, as detailed in Table~\ref{tab:Subjects}. 
Note that, to avoid any selection bias, we reused the training and test sets defined in the original sources of these datasets~\cite{krizhevsky2009learning, ILSVRC15}. In our experiments, we use the training set to both construct an RCS and train a linear aligner for each subject, as explained in Section~\ref{sec:Background}. We then use the test sets for fault detection and evaluation, as detailed in the next section.

\subsection{Concept extraction using VLMs}

As explained in Section~\ref{sec:Approach}, we use a concept-based feature for constructing CAFD. To do this, we first need to extract semantic concepts from image inputs.
We rely on the concept extraction approach from our prior work~\cite{abbasishahkoo2026highly}, as described in Section~\ref{sec:Background}. This approach consists of two main steps. First, an RCS is constructed from the MUT's training set by extracting its concepts using the CLIP model~\cite{radford2021learning}. Second, concepts representing each unlabeled test input are extracted using the constructed RCS. In addition, a linear aligner for mapping the representation of image inputs from the MUT's input space to the CLIP's input space, constructed based on the MUT's training set, is used in this step to avoid the need to execute CLIP on every test input and further improve the efficiency of our approach.
In our experiments, we used the original implementation along with the VLM model and configurations proposed by Abbasishahkoo \textit{et al.}~\cite{abbasishahkoo2026highly}. 

To ensure the reliability of this mapping, we evaluated the learned aligner's accuracy using the coefficient of determination ($R^2$). The achieved accuracies, 0.83 for Cifar-10 (ResNet18), 0.82 for Cifar-100 (ResNet152),  and 0.72 for ImageNet (ResNet101), are consistent with those reported in prior work~\cite{abbasishahkoo2026highly,moayeri2023text}, indicating that the aligner achieves a level of accuracy preserving sufficient semantic information from the original CLIP space while avoiding the computational overhead of executing CLIP on every test input. Consequently, this provides confidence in the reliability of the extracted concepts for calculating our concept-based feature and CFR scores, as well as for training the CAFD model. 

\begin{table*}[b]
    \centering   
    \footnotesize 
    \caption{FDR achieved by the proposed approach and baselines across different subjects and budget sizes}
    \resizebox{0.6\textwidth}{!}{
    \begin{tabular}{|c|c|      c|  c|  c| c|  c| c|  }
    \hline 
    \multirow{2}{*}{Dataset} &Budget &\multirow{2}{*}{CAFD} &\multicolumn{5}{c|}{Baselines} \\ \cline{4-8}
       &size      &   &DeepGini  &Margin   &Vanilla &DATIS &RTS  \\ \hline

    \multirow{6}{*}{Cifar-10}  
    &$b$=1\%   &\textbf{0.75}       &0.61       &0.55      &0.55         &\underline{0.68}  &0.59 \\ \cline{2-8}
    &$b$=3\%   &\textbf{0.57}		&0.49       &0.41      &0.44         &\underline{0.56}  &0.43 \\ \cline{2-8}
    &$b$=5\%   &\textbf{0.73}		&0.59       &0.57      &0.59         &\underline{0.71}  &0.59 \\ \cline{2-8}
    &$b$=7\%   &\textbf{0.85}		&0.73       &0.72      &0.73         &\underline{0.82}  &0.71 \\ \cline{2-8}
    &$b$=10\%  &\textbf{0.96}		&0.87       &0.86      &0.87         &\underline{0.92}  &0.86 \\ \cline{2-8}
    &$b$=12\%  &\textbf{0.97}		&0.92       &0.92      &0.92         &\underline{0.94}  &0.91 \\ \hline

    \multirow{6}{*}{Cifar-100}  
    &$b$=1\%   &\textbf{0.78}       &\underline{0.68}       &0.48      &0.65         &\underline{0.68}  &0.63 \\ \cline{2-8}
    &$b$=3\%   &\textbf{0.64}		&\underline{0.60}       &0.46      &0.60         &\underline{0.60}  &0.51 \\ \cline{2-8}
    &$b$=5\%   &\textbf{0.58}		&\underline{0.56}       &0.45      &0.54         &0.52  &0.49 \\ \cline{2-8}
    &$b$=7\%   &\textbf{0.55}		&\underline{0.52}       &0.45      &0.50         &0.47 &0.46 \\ \cline{2-8}
    &$b$=10\%  &\textbf{0.71}		&\underline{0.63}       &0.59      &0.63         &0.59  &0.57 \\ \cline{2-8}
    &$b$=12\%  &\textbf{0.74}		&0.69       &0.67      &\underline{0.71}         &0.66  &0.63 \\ \hline

    \multirow{6}{*}{ImageNet}  
    &$b$=1\%   &\textbf{0.81}       &\underline{0.76}       &0.55      &0.74         &0.75  &0.66 \\ \cline{2-8}
    &$b$=3\%   &\textbf{0.74}		&\underline{0.67}       &0.48      &0.65         &0.66  &0.56 \\ \cline{2-8}
    &$b$=5\%   &\textbf{0.69}		&\underline{0.63}       &0.51      &0.60         &0.62  &0.55 \\ \cline{2-8}
    &$b$=7\%   &\textbf{0.65}		&\underline{0.59}       &0.51      &0.57         &0.58  &0.52 \\ \cline{2-8}
    &$b$=10\%  &\textbf{0.60}		&\underline{0.53}       &0.48      &0.52         &0.52  &0.49 \\ \cline{2-8}
    &$b$=12\%  &\textbf{0.56}		&\underline{0.51}       &0.48      &\underline{0.51}         &\underline{0.51}  &0.47 \\ \hline    
    
    \end{tabular}
    }
    \label{tab:FDR_results}
\end{table*}

\subsection{Evaluation metric}
To assess the effectiveness of CAFD and compare its performance against baselines in terms of fault detection capability, we use the FDR metric. FDR is a well-established measure in the DNN testing literature~\cite{aghababaeyan2024deepgd, li2024distance, abbasishahkoo2024teasma} that quantifies the effectiveness of test selection methods in identifying faults revealed by the selected test subset under a defined selection budget. The FDR value for a test subset $T$ containing the top $b$ inputs of an ordered list produced by each approach is computed as follows:

\begin{equation}\label{Eq:FDR}
FDR(T)= \frac{\left|F_{T}\right|}{\min(b, \left|F\right|)}
\end{equation}

where $\left|F_{T}\right|$ is the number of distinct faults revealed by $T$, $\left|F\right|$ is the total number of faults, and and $b$ denotes the selection budget, i.e., size of $T$.
To estimate faults in the MUT, we rely on the approach proposed by Aghababaeyan \textit{et al.}~\cite{aghababaeyan2023black}, in which faults are defined as distinct root causes of failures. As outlined in Section~\ref{sec:Background}, this approach consists of three main steps: feature extraction, dimensionality reduction, and density-based clustering. In our experiments, we used the original implementation and parameter settings provided by Aghababaeyan \textit{et al.}. Table~\ref{tab:Subjects} reports the total number of faults identified for each subject.

\par\vspace{-7pt}
\subsection{Data Availability}
The replication package for our study will be shared upon the paper's acceptance~\cite{replicationpackage}.


\section{Results}
\label{sec:Results}

In this section, we present the results related to our research questions and discuss their practical implications. 


\subsection{\textbf{RQ1: Effectiveness of CAFD}}

To address this research question, we applied CAFD alongside all baseline methods described in Section~\ref{sec:Background} to rank unlabeled test inputs. We then evaluate the number of distinct faults detected by test subsets of varying budget sizes, selected from the top of each ranked list. 
We conducted our evaluations under highly constrained labeling budgets, aligning with the practical limitations imposed by the cost and effort of labeling inputs. We computed \textit{FDR} for various budget sizes $b$ 1\%, 3\%, 5\%, 7\%, 10\%, and 12\% of each test set. Furthermore, to ensure a fair comparison, we used the same configurations and parameter settings provided in the original implementation of each baseline

Table~\ref{tab:FDR_results} provides the FDR results across all subjects and budget sizes. The highest FDR for each subject and selection budget in Table~\ref{tab:FDR_results} is highlighted in bold, and the second-highest FDR is underlined. The results demonstrate that CAFD consistently outperforms all baselines across subjects and selection budgets. The improvements are more significant at smaller budget sizes, highlighting the effectiveness of our method in prioritizing highly fault-revealing inputs. For example, at a budget size of 1\%, CAFD achieves improvement percentages of 22\%, 31\%, and 21\% on Cifar-10, Cifar-100, and ImageNet, respectively, compared to the second-best method\footnote{Improvement percentage =  $\frac{FDR_{\text{our}} - FDR_{\text{second-best}}}{1 - FDR_{\text{second-best}}} \times 100$.}.
These improvements remain consistent across other constrained budgets, where CAFD demonstrates robust overall performance, with average improvements of 10\% on Cifar-10, 15\% on Cifar-100, and 19\% on ImageNet across highly constrained budgets (1\%, 3\%, and 5\%).

It is important to note that the second-best performing approach varies across subjects and budget sizes. On Cifar-100, DeepGini, Vanilla, and DATIS are, in turn, the second-best baselines depending on the budget size. On ImageNet, DeepGini is the second-best-performing baseline for smaller budgets. However, as the budget increases, the differences in FDR between DeepGini, Vanilla, and DATIS become negligible, typically around 0.01. In contrast, on Cifar-10, DATIS remains the second-best baseline across all budget sizes.
This variation highlights the robustness of CAFD, making it the only method we can confidently rely on across DNNs, datasets, and budget sizes. 


The improvements provided by CAFD in FDR relative to the second-best baseline are substantial, particularly on ImageNet.
To further validate these results, we conducted Wilcoxon signed-rank tests comparing CAFD with each baseline at a significance level of $\alpha = 0.05$. The results show that all p-values are below 0.05, confirming that the observed differences are statistically significant. These findings indicate that CAFD consistently outperforms all SOTA baselines in terms of fault-revealing power.

\begin{tcolorbox}
    \textbf{Answer to RQ1.} CAFD consistently outperforms all baselines in detecting distinct faults under the same labeling budget. Moreover, it achieves statistically and practically significant improvements, particularly under highly constrained budgets, with average improvement percentages of 10\%, 15\%, and 19\% on Cifar-10, Cifar-100, and ImageNet, respectively, at budget sizes of 1\%, 3\%, and 5\%, compared to the second-best approach. Furthermore, the second-best approach is not consistent across subjects and budget sizes, making CAFD the only reliable approach that consistently delivers superior performance.
\end{tcolorbox}

\subsection{\textbf{RQ2: Efficiency of CAFD}}
To answer this research question, we empirically evaluate the execution time of CAFD and compare it against SOTA baselines. Specifically, we measure the time required by each approach to estimate the fault-revealing probability for all inputs in a test set. It is important to note that, similar to RTS, CAFD includes an initial setup phase. As described in Section~\ref{sec:Approach}, this phase primarily consists of calculating the values for the selected features for each training input, and then training the CAFD LR model. 
However, this initial phase must be performed only once for each MUT. Once trained, CAFD can estimate the fault-revealing probabilities for any number of unlabeled test inputs, making it even more cost-effective for larger test sets. Therefore, similar to RTS, we exclude the cost of this one-time setup phase from the execution time measurements reported for CAFD and RTS.
Table~\ref{tab:time} reports the execution time (in seconds) of CAFD and each baseline for estimating the fault-revealing probability of the entire test set for each subject.

\begin{table}[b]
    \centering   
    \caption{Execution time (in seconds) of CAFD and each baseline on the entire test set.}
    \resizebox{\columnwidth}{!}{
    \begin{tabular}{   |cc|    c|    c|c|c|c|c|  }
    \cline{1-8} 
     \multirow{2}{*}{Model} &\multirow{2}{*}{Dataset}   &\multirow{2}{*}{CAFD} &\multicolumn{5}{c|}{Baselines}  \\ \cline{4-8}
      &          &   &DeepGini &Margin &Vanilla   &DATIS &RTS \\ \cline{1-8}

    ResNet18 &Cifar-10  &8.43  &3.502  &3.52  &3.503  &8.50  &141     \\ \cline{1-8}
    ResNet152 &Cifar-100  &35.99 &29.01  &29.02  &29.02  &137  &1200     \\ \cline{1-8}
    ResNet101 &ImageNet   &839.8  &105.13  &105.29  &105.02  &1321  &104400    \\ \cline{1-8}
                      
    \end{tabular}    
     \label{tab:time}
     }
\end{table}

As shown in Table~\ref{tab:time}, although CAFD generally requires more time than simpler baselines, such as probability-based uncertainty metrics, it is not the most computationally expensive approach and remains a highly practical solution. This is particularly important for complex DNNs and large datasets with many output classes, such as ResNet-101 with the ImageNet dataset. For this subject, CAFD requires approximately 839 seconds to process the entire ImageNet test set with 50,000 inputs. In contrast, while RTS remains practical for smaller subjects, it does not scale effectively and becomes infeasible for large-scale settings such as ImageNet.
Moreover, since fault detection in a DNN model is neither frequent nor performed in real time, moderate increases in execution time are acceptable when they lead to improved fault detection effectiveness.

It is important to note that, to ensure a fair and consistent comparison, all execution times reported in this table were measured on a server equipped with an Intel(R) Xeon(R) Gold 6234 CPU (3.30 GHz) and an Nvidia Quadro RTX 6000 GPU with 24 GB of memory. However, CAFD’s execution time can be further reduced in industrial computing environments, which are typically more powerful and support enhanced parallelization. This further supports the practicality of CAFD.

Given our earlier findings, where CAFD consistently outperforms all baselines and achieves practically significant improvements in fault detection, particularly under highly constrained budgets, its execution time remains acceptable even for complex DNNs such as ResNet101 and large datasets like ImageNet with 1000 classes. CAFD is thus a cost-effective alternative, as it not only delivers significant improvements in FDR but also remains computationally efficient, even for deep and complex networks.

\begin{tcolorbox}
    \textbf{Answer to RQ2.} Although CAFD’s computational cost is higher than that of simple uncertainty-based baselines, it remains highly practical and scalable. Our results demonstrate this even for subjects with large datasets with many output classes, and deep, complex DNN architectures. Given its superior FDR performance, CAFD offers a strong trade-off between effectiveness and efficiency, making it a cost-effective solution for DNN fault detection.
\end{tcolorbox}  

\subsection{Threats to Validity}
In this section, we discuss the different potential threats to the validity of our study and describe how we mitigated them.

\textbf{Construct threats to validity.}
A potential threat arises from the implementation of fault detection baselines. To minimize the risk of implementation bias and ensure a fair comparison, we rely on the original implementations provided by the respective authors.  
Another construct threat to validity concerns the estimation of faults in DNNs, which is inherently more challenging than in traditional software systems. Consequently, identifying and quantifying faults may be subject to approximation errors. To mitigate this threat, we adopt a well-established clustering-based fault estimation approach that has been widely used in prior studies~\cite{li2024distance, aghababaeyan2024deepgd, abbasishahkoo2024teasma}, thereby ensuring consistency with existing evaluation practices.


\textbf{Internal threats to validity.}
One potential internal threat to validity arises from the selection of parameters and experimental configurations for both the baseline approaches and CAFD. To mitigate this threat, we adopt the default settings and configurations provided in the original implementation of all baselines, ensuring a fair and consistent comparison. 
Another source of internal validity concerns the dependence of both the CFR metric and CAFD on the chosen VLM and the method used to extract textual concepts from image inputs. As described in Section~\ref{sec:concept_extraction}, we employ CLIP together with an efficient procedure for mapping the unlabeled images' representation in the MUT to CLIP's shared space and calculating their similarity with textual concepts' embeddings. Moreover, both CAFD and the CFR metric rely on the parameter $m$, which determines the number of highly similar concepts extracted from each input. Following prior work~\cite{abbasishahkoo2026highly}, we use the default setting of $m=10$. While this choice ensures consistency with established practices, it may not be optimal for all settings.
It is important to note that using more advanced VLMs, alternative concept extraction techniques, or tuning the parameter $m$ may further improve the quality of the extracted concepts, potentially enhancing the effectiveness of CFR and, consequently, the performance of CAFD.
 


\textbf{External threats to validity.}
A potential threat to external validity concerns the generalizability of CAFD across different models, datasets, and usage scenarios. To mitigate this threat, we conducted experiments on multiple subjects encompassing diverse DNN architectures and input datasets with varying characteristics. In particular, we included large-scale benchmarks such as ImageNet to evaluate both scalability and practical efficiency in realistic settings. Furthermore, we considered a range of selection budgets to assess CAFD's adaptability under different resource constraints and compared its performance against multiple baselines. Across all combinations of datasets, model architectures, and budget sizes, CAFD consistently achieved high FDR performance. 
Moreover, our empirical evaluation is conducted on image inputs because we use VLMs. While CAFD is not inherently restricted to images and can generalize to other modalities by replacing the underlying VLM with modality-specific encoders, further evaluation on non-visual domains is part of our future work.
These results provide strong evidence of its robustness and generalizability across diverse usage scenarios.

\section{Related Work}
In this section, we first review existing work on DNN fault detection approaches. We then briefly summarize prior work that leverages VLMs to evaluate and improve DNNs.

\subsection{DNN Fault Detection}
In recent years, several approaches have been proposed for DNN fault detection. However, many earlier approaches have shown limited effectiveness and have been consistently outperformed by more recent techniques. For instance, early contributions such as NC~\cite{pei2017deepxplore, Ma2018DeepGaugeMT}, Cross Entropy-based Sampling (CES)~\cite{li2019boosting}, SA~\cite{kim2019guiding}, and ATS~\cite{gao2022adaptive} have been extensively studied in prior work. NC metrics prioritize inputs based on the extent to which neuron activation patterns are exercised. In contrast, SA metrics quantify how surprising an unlabeled input is with respect to the MUT’s training input distribution. CES prioritizes inputs using representations learned by the MUT, where each input is characterized by the probability distribution of neuron outputs in the last hidden layer, and selection is guided by minimizing cross-entropy.
ATS iteratively selects inputs based on the dissimilarity between each remaining input and the currently selected subset, making it one of the most computationally intensive approaches~\cite{hu2024test}. 
Despite their important contributions, these approaches have been consistently outperformed by more recent fault detection methods, including several baselines considered in our experiments, such as RTS~\cite{sun2023robust} and DATIS~\cite{li2024distance}.

Uncertainty-based approaches have become one of the most prominent categories of DNN fault detection. The main objective of these approaches is to prioritize inputs for which the MUT exhibits low confidence in its predictions~\cite{feng2020deepgini, hu2024test, Weiss2022SimpleTechniques}. As a result, they tend to select inputs located near the MUT’s decision boundary, where the MUT failures are more likely to occur. DeepGini~\cite{feng2020deepgini}, Vanilla Softmax~\cite{Weiss2022SimpleTechniques}, and Margin~\cite{hu2024test} are among the most widely used probability-based uncertainty metrics and rely solely on the MUT’s predicted probabilities. 
In addition to their computational efficiency, these methods have demonstrated strong effectiveness in DNN fault detection across various studies~\cite{hu2024test, Weiss2022SimpleTechniques, feng2020deepgini, li2024distance}. 
Beyond probability-based uncertainty metrics, recent studies have explored leveraging information from neighboring inputs to improve uncertainty estimation for DNN fault detection~\cite{li2024distance, bao2023defense}. DATIS~\cite{li2024distance} 
estimates uncertainty by incorporating the distance between the given input and its nearest neighbors in the MUT’s labeled training set. 
While both probability-based and neighbor-aware uncertainty metrics frequently rank second among baselines in our results, CAFD consistently outperformed all three methods across all subjects and selection sizes.


Hybrid and learning-based approaches are also among the well-studied and effective fault-detection approaches for DNNs. These approaches aim to leverage the complementary strengths of multiple information sources to enhance the effectiveness of DNN fault detection~\cite{aghababaeyan2024deepgd, sun2023robust, li2021testrank, demir2024test}. 
However, combining multiple metrics often introduces additional computational cost. In some cases, this cost can become significant, reducing the practicality and scalability of such approaches~\cite{aghababaeyan2024deepgd, abbasishahkoo2025metasel}.
RTS proposed by Sun \textit{et al.}~\cite{sun2023robust} is a hybrid approach that integrates uncertainty and diversity for DNN fault detection.
As shown in our results, RTS is the most time-consuming approach, taking approximately 17 to 117 times longer than CAFD to prioritize the entire test set across our subjects. Moreover, RTS consistently underperforms CAFD in effectiveness and never ranks among the second-best baselines.
DeepGD~\cite{aghababaeyan2024deepgd} is another hybrid search-based fault detection approach that combines uncertainty and geometric diversity and leverages the NSGA-II algorithm~\cite{deb2002fast}. However, a recent study~\cite{abbasishahkoo2026highly} highlights notable scalability limitations, reporting a substantial decline in efficiency on large-scale datasets. In particular, their results on ImageNet indicate that executing DeepGD requires more than three times the runtime of RTS.

Mutation-based approaches~\cite{wang2021prioritizing, hu2019deepmutation++, abbasishahkoo2024teasma} select inputs based on the number of mutants they kill, motivated by the intuition that such inputs are more likely to reveal DNN failures. However, these approaches are computationally expensive and impractical, as they require generating many mutants and evaluating each unlabeled test input across all generated mutants~\cite{abbasishahkoo2024teasma, wang2021prioritizing}.


\subsection{DNN Testing Using VLMs}
The emergence of VLMs has opened new opportunities for analyzing and improving DNN models. In a recent study, Hu \textit{et al.}~\cite{hu2025debugging} employ the CLIP model to perform runtime analysis of image classification DNNs and to enable fault localization within their network components. The core of their approach lies in semantic heatmaps that capture how frequently a given concept appears in correctly and incorrectly classified training inputs. These heatmaps are then leveraged at runtime to assess unlabeled inputs. By measuring the similarity between each input and all heatmaps, they can predict whether the model is likely to fail in predicting the input and, ultimately, estimate model accuracy. In addition, through zero-shot classification with a VLM, they distinguish whether failures originate in the encoder or the classification head of the DNN. While their approach similarly leverages the CLIP model for DNN analysis, its objective differs from our fault-detection goal. Furthermore, the size of these heatmaps poses a key limitation for the scalability and generalizability of their approach. To control this complexity, they restrict the approach to a small, predefined set of concepts.
However, applying their approach to datasets with a large number of classes, such as ImageNet with 1000 classes, requires a substantially larger set of concepts, leading to prohibitively large heatmaps and limiting practical applicability. In contrast, CAFD relies on CFR, which uses a flexible, diverse set of concepts. Moreover, CFR is computed for each concept by quantifying its frequency in failure-causing inputs, avoiding the need for semantic heatmaps and resulting in a more computationally efficient and scalable solution.
Abbasishahkoo et al.~\cite{abbasishahkoo2026highly} also leveraged CLIP for DNN Improvement through retraining. They propose a concept-based diversity metric to select a diverse subset of unlabeled inputs for labeling and subsequent DNN retraining. Instead of manually specifying concepts, they rely on a knowledge base to extract a diverse set of highly related concepts. While their objective focuses on improving model performance via retraining, which differs from our fault detection goal, we leveraged their strategy for concept extraction, as described in Section~\ref{sec:concept_extraction}.

\section{Conclusion}

In this paper, we introduce CAFD, a hybrid, learning-based fault-detection approach for DNNs. To construct CAFD, we investigate various types of models, including both classical machine learning techniques and NNs, and carefully select a set of effective features to yield a fault detection model that is both highly accurate and practically efficient. In particular, we combine features derived from the MUT's output---such as uncertainty metrics---with a distance-based feature that considers an input's nearest neighbors and a novel concept-based feature extracted from image inputs using VLMs. We introduce CFR, a metric that measures how frequently a given textual concept appears in failure-causing image inputs within the MUT’s training set. To extract these textual concepts from image inputs, we leverage VLMs. We conducted an extensive empirical evaluation of CAFD, comparing it against five SOTA baselines across three subject DNN models and datasets, including the large ImageNet dataset. Across a wide range of constrained selection budgets, our results show that CAFD consistently outperforms all baselines in terms of FDR. Moreover, the second-best approach varies across datasets and budget sizes, highlighting CAFD as the only consistently reliable approach. In addition to its superior performance, our evaluation results show that CAFD achieves strong practical efficiency even on large-scale datasets like ImageNet.

%

\section*{Acknowledgements}
This work was supported by the Canada Research Chair and Discovery Grant programs of the Natural Sciences and Engineering Research Council of Canada (NSERC) and the Research Ireland grant 13/RC/2094-2.


\bibliographystyle{IEEEtran}
\bibliography{main.bib}

\begin{thebibliography}{10}
\bibitem{replicationpackage}
The link to these artifacts will be made publicly available upon acceptance of this paper.
\end{thebibliography}

\end{document}